\documentclass[11pt,a4paper]{article}
\pdfoutput=1
\usepackage[hyperref]{emnlp2018}
\usepackage{times}
\usepackage{latexsym}

\usepackage{url}
\usepackage{graphicx}
\usepackage{tabularx}
\usepackage{multirow}

\aclfinalcopy % Uncomment this line for the final submission

\title{Comparing CNN and LSTM character-level embeddings in  BiLSTM-CRF models for chemical and disease named entity recognition}

\author{Zenan Zhai, Dat Quoc Nguyen \and Karin Verspoor  \\
School of Computing and Information Systems \\
The University of Melbourne, Australia \\
{\tt {\small zenanz@student.unimelb.edu.au, \{dqnguyen, karin.verspoor\}@unimelb.edu.au}}
}

\begin{document}
\maketitle
\begin{abstract}
We compare the use of LSTM-based and CNN-based character-level word embeddings in BiLSTM-CRF models to approach chemical and disease named entity recognition (NER) tasks.  Empirical results over the BioCreative V CDR corpus show that the use of either type of character-level word embeddings in conjunction with the BiLSTM-CRF models leads to comparable state-of-the-art performance. However, the models using CNN-based character-level word embeddings have a computational performance advantage, increasing training time over word-based models by 25\% while the LSTM-based  character-level word embeddings more than double the required training time.
\end{abstract}

\section{Introduction}
Bi-directional Long-Short Term Memory Conditional Random Field models (BiLSTM-CRF), in which a BiLSTM is coupled with a CRF layer to connect output tags, have been shown to achieve state-of-art performance in sequence tagging tasks including  part of speech (POS) tagging, chunking, and NER \cite{Huang:2015}. 
%In prior research, 
The combination of word embeddings and character-level word embeddings has been explored in this context, with \newcite{Ma:2016} using Convolutional Neural Networks (CNNs) to construct  character-level word embeddings and \newcite{Lample:2016} applying LSTM networks.
%for learning the character-level word embeddings.
This work showed that the use of character-level word embeddings improves the performance of the models, by contributing the ability to recognize unseen words.

Biomedical Named Entity Recognition (BNER) is a vital initial step for information extraction tasks in the biomedical domain, including the Chemical-Disease Relationship (CDR) extraction task where both chemical and disease entities must be identified \cite{Li:2016}. 
Character-level word embeddings could be particularly significant in this context, given that new entity names are frequently created, and may follow consistent patterns including productive morphology such as common prefixes (e.g., \textit{di-}) or suffixes (e.g., \textit{-ase}). Features that capture word-internal characteristics have been shown to be effective for BNER tasks in CRF models  \cite{klinger:2008}.

\newcite{Lyu:2017} applied a BiLSTM-CRF model with LSTM-based character-level word embeddings to a gene and protein NER task,
%\newcite{Lyu:2017}  demonstrated that the BiLSTM-CRF model with LSTM-based character-level embeddings 
demonstrating state-of-art performance that outperformed traditional feature-based models. 
 \newcite{Luo:2018} further improved on this result on a chemical NER task by adding an attention layer between the BiLSTM and CRF layers (Att-BiLSTM-CRF).

In an experiment by \newcite{Nils:2017}, optimal hyper-parameters for LSTM networks in sequence tagging tasks were explored, with the finding that incorporation of character-level word embeddings  significantly improved performance on NER tasks on general datasets including CoNLL 2003 \cite{Sang:2003}. However, the choice of CNN-based \cite{Ma:2016}  or LSTM-based character-level word embeddings \cite{Lample:2016} did not affect the performance significantly. 
Since the CNN has fewer parameters to train than BiLSTM network, it is better in terms of training efficiency, and was recommended as the preferred approach. 

%However, only BiLSTM-CRF models with LSTM embedding are tested on the BNER task so far. Therefore, in this paper,
In this paper, we implement and compare models with each type of word embedding to generate empirical results for the tasks of chemical and disease NER, using the BioCreative V CDR corpus \cite{Li:2016}. These BNER categories are the most searched entities in the biomedical literature \cite{Dogan:2009}, and hence particularly important to study. 

The results show that models with CNN-based character-level word embeddings achieve state-of-the-art results comparable to LSTM-based character-level word embeddings, while having the advantage of reduced training complexity, demonstrating that the prior results also hold for the BNER task.

\begin{figure*}[!t]
\centering
\includegraphics[width=14.75cm]{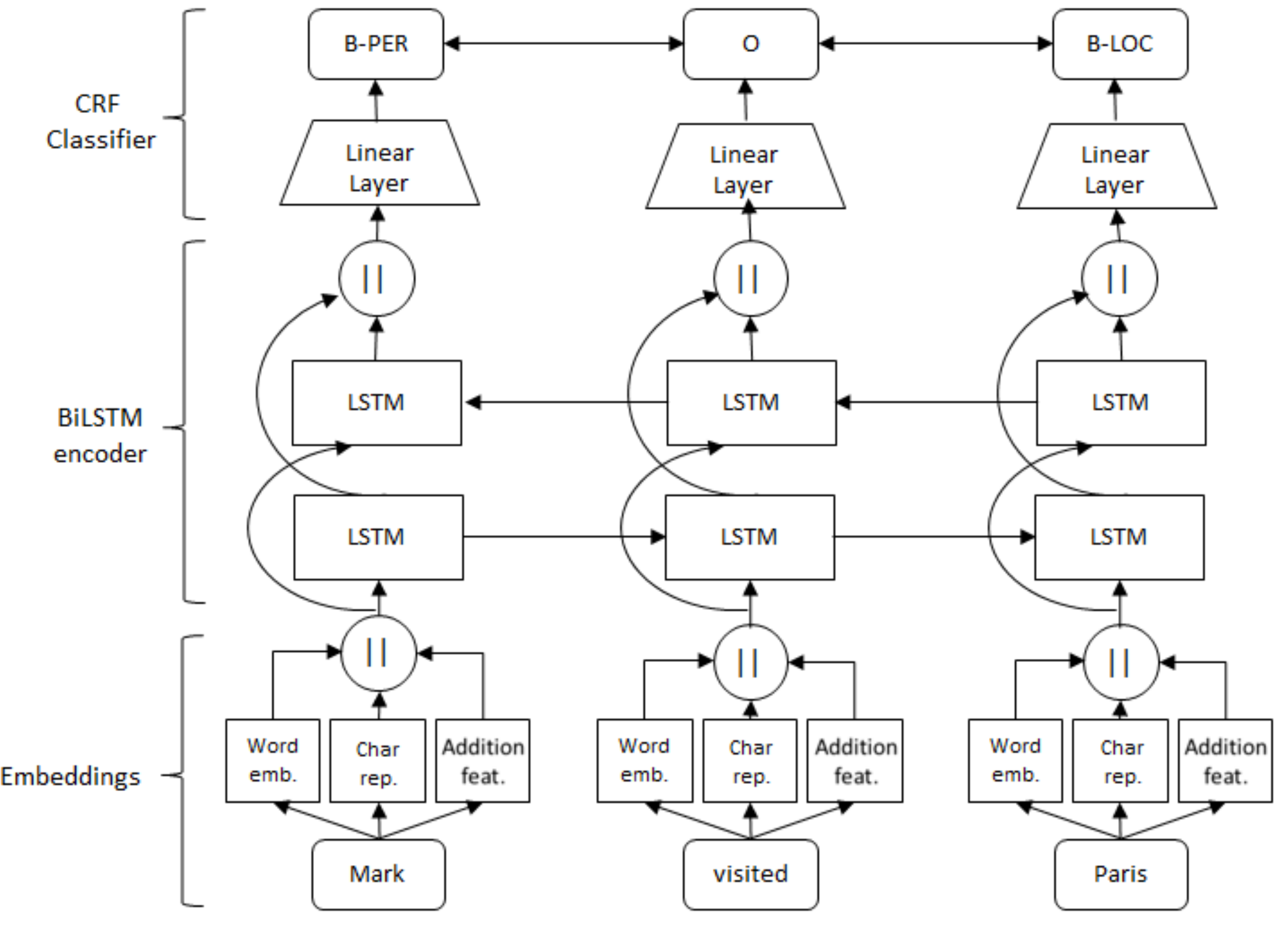}
\caption{Architecture of BiLSTM-CRF models with character-level word representations and additional features. This figure is adapted from  \newcite{ReimersG17}.}
\label{fig:model}
\end{figure*}

\begin{figure*}[!t]
\centering
$\begin{array}{cc}
\includegraphics[width=7.75cm]{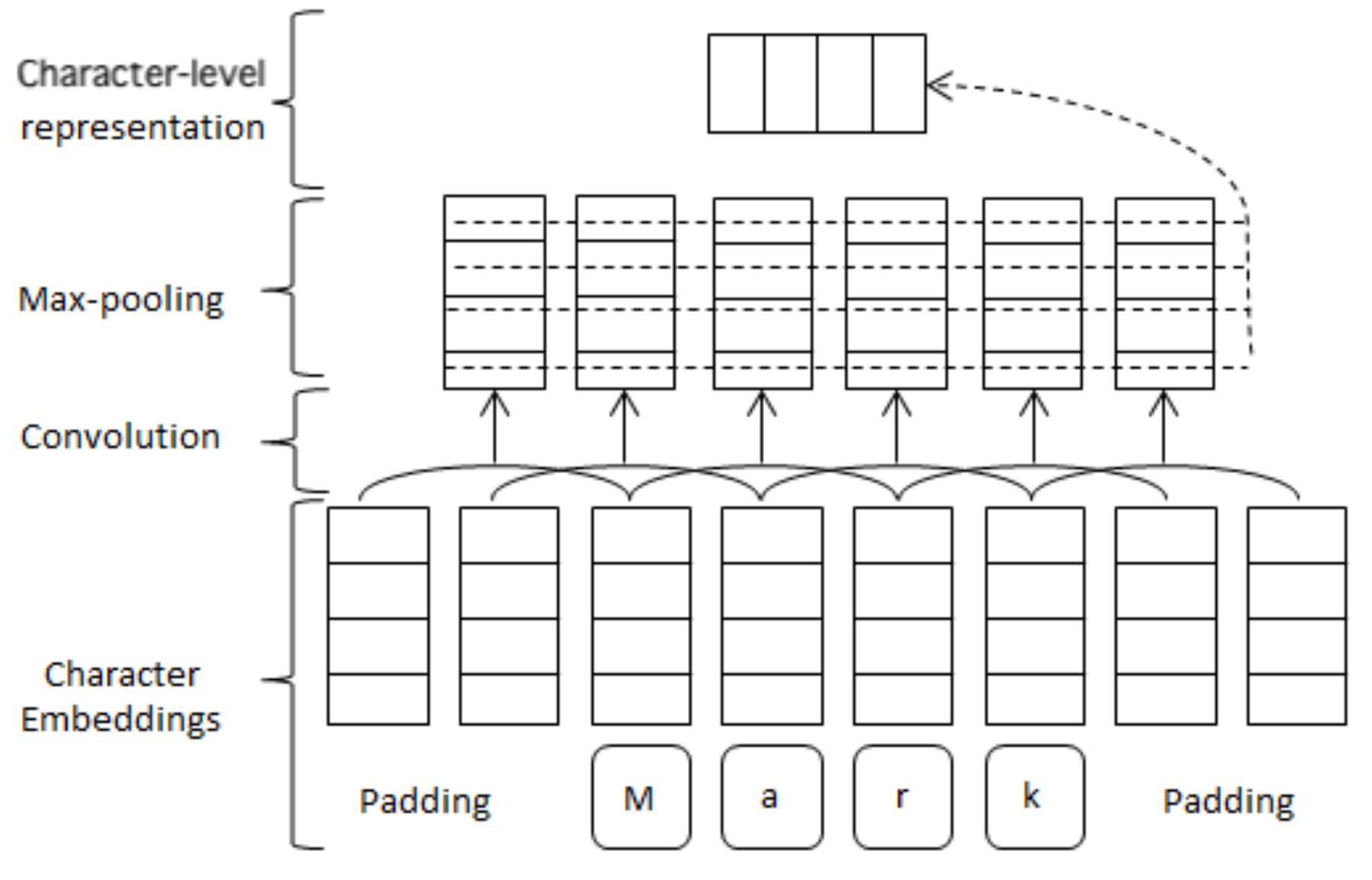} &
\includegraphics[width=7cm]{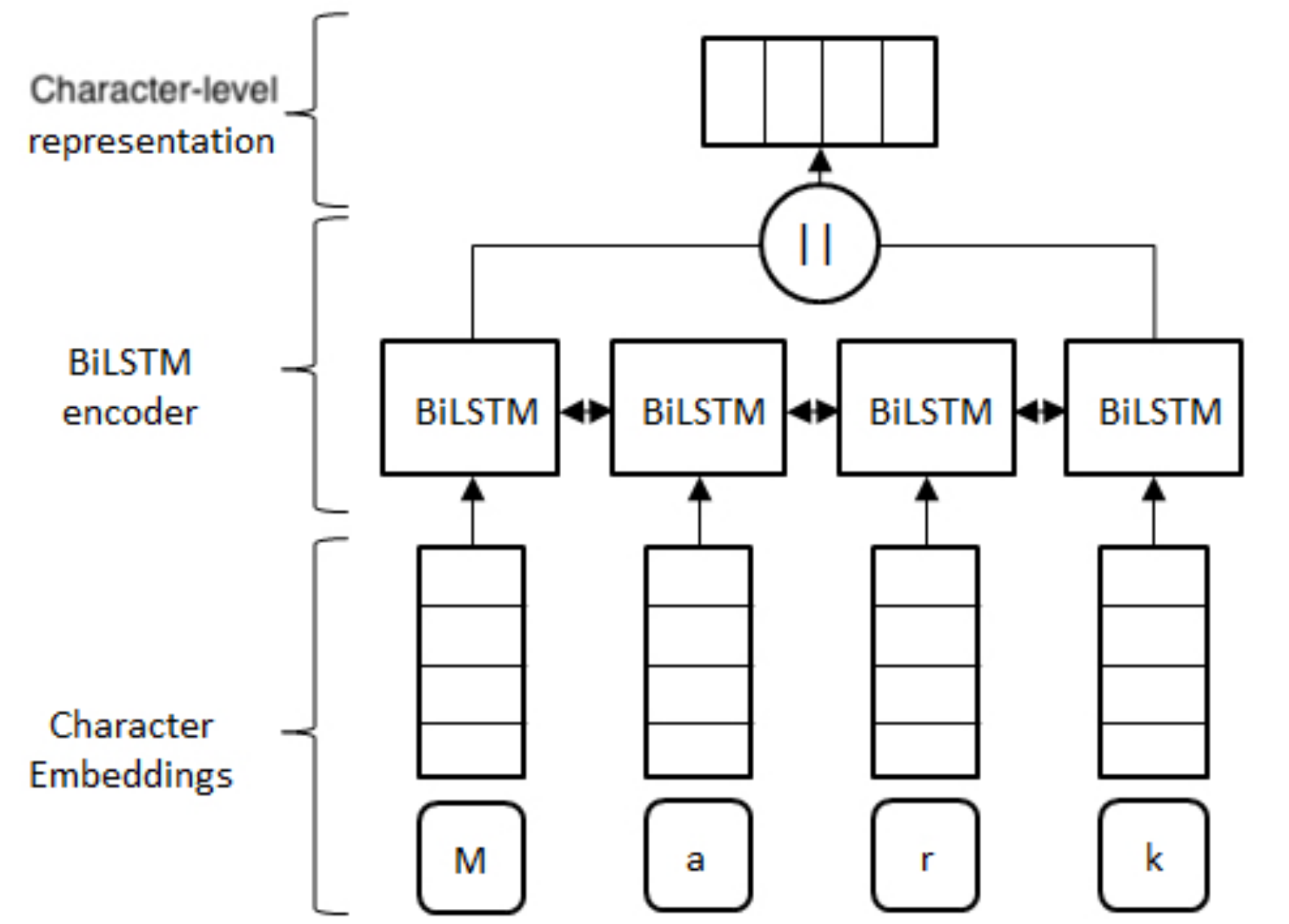} 
\\
\mbox{({CNN-based character-level word representation})} & \mbox{({LSTM-based character-level word representation})}
\end{array}$
\caption{Character-level word representations. This figure is also adapted from  \newcite{ReimersG17}.}
\label{fig:characters}
\end{figure*}

\section{Experimental methodology}

This section presents our empirical approach to comparing state-of-the-art neural network models for chemical and disease NER.

\subsection{Dataset}

In our experiments, we use the BioCreative V  CDR corpus \cite{Li:2016}. This corpus provides a set of 1000 manually-annotated abstracts ({9193} sentences) for training and development, and another set of 500 manually-annotated abstracts ({4840} sentences) for test. In particular, we used a  pre-processed version of the CDR corpus from  \newcite{Luo:2018},\footnote{\url{https://github.com/lingluodlut/Att-ChemdNER}} which provides predicted POS-, chunking-  and gazetteer-based tags:

\begin{itemize}
    \item POS and chunking tags are predicted by  the GENIA tagger \cite{Yoshimasa:2005}.\footnote{\url{http://www.nactem.ac.uk/GENIA/tagger}}% and incorporated as additional features to explore the effect of syntactic features on the BiLSTM-CRF models. 
    \item Gazetteer tags are encoded in BIO tagging scheme based on matching to the external Jochem chemical dictionary \cite{Hettne:2009}. %  this dictionary only contains chemical entities. 
\end{itemize}

Following \newcite{Luo:2018}, we randomly sample 10\% from the set of 1000 abstracts for development, and use the remaining for training.

\subsection{Models}
We use the following  BiLSTM-CRF-based sequence labeling models: %  for sequence labeling:
\begin{itemize}
    \item Baseline BiLSTM model \cite{Schuster1997BRN,HochreiterSchmidhuber1997b} which uses a softmax layer to predict NER labels of input words.

    \item BiLSTM-CRF \cite{Huang:2015} extends the BiLSTM model with a CRF layer which allows the model to use sentence-level tag information for sequence prediction. 
    
    \item BiLSTM-CRF + CNN-char %, i.e.\ BiLSTM-CNN-CRF
    \cite{Ma:2016}  extends the BiLSTM-CRF model with character-level word embeddings. For each word, its  character-level word embedding is derived by applying a CNN to the character sequence in the word. 
    
    \item BiLSTM-CRF + LSTM-char %, or BiLSTM-LSTM-CRF,
    also extends the BiLSTM-CRF model with character-level word embeddings which are derived by applying a BiLSTM to the character sequence in each word  \cite{Lample:2016}.%: forward and backward LSTMs  are applied on the character sequence of each word in forward and backward directions, respectively; and then the character-level word representations were obtained by concatenating the outputs from both directions.
\end{itemize}

Following \newcite{Luo:2018}, we also consider the impact of extra features including  syntactic features such as POS and chunking tags, 
and a chemical term feature based on matching to an external gazetteer. Figure \ref{fig:model} illustrates the general BiLSTM-CRF model architecture  with character-level word embeddings and additional features, while Figure \ref{fig:characters} illustrates CNN-based and LSTM-based architectures for learning the character-level word embeddings.

\begin{table}[t!]
\begin{center}
\begin{tabular}{|l|l|}
\hline \bf Hyper-para. & \bf Value \\ \hline
Optimizer & Nadam \\
 Mini-batch size	 & 32 \\
 Clipping & $\tau = 1$ \\
 Dropout & [0.25, 0.25] \\
% Backend & Theano \\
 \hline
\end{tabular}
\end{center}
\caption{\label{config:default} Fixed hyper-parameter configurations. }
\end{table}

\begin{table}[t!]
\centering
\resizebox{7.75cm}{!}{
\begin{tabular}{cc}
CNN-based & LSTM-based\\
\begin{tabular}{|l|l|}
\hline
\textbf{Hyper-para.} & \textbf{Value}\\\hline
charEmbedSize & 30 \\
Window size & 3 \\
\# of filters & 30 \\
%Output size & 30 \\
\hline
\# of Params. & 2,730 \\\hline
\end{tabular} &
\begin{tabular}{|l|l|}
\hline
\textbf{Hyper-para.} & \textbf{Value}\\\hline
charEmbedSize & 30 \\ 
BiLSTM layer & 1 \\ 
LSTM size & 25 \\
%Output size & 50\\
\hline
\# of Params. & 11,200 \\\hline
\end{tabular}
\end{tabular}
}
\caption{Hyper-parameters for learning character-level word embedding. ``charEmbedSize'' and ``\# of Params.'' denote the vector size of character embeddings and  the total number of parameters, respectively.}
\label{config:embedding}
\end{table}

\subsection{Implementation details}

We used  a well-known implementation of BiLSTM-CRF-based  models from \newcite{Nils:2017}.\footnote{\url{https://github.com/UKPLab/emnlp2017-bilstm-cnn-crf}} We used the training set to learn model parameters,
the development set to select optimal  hyper-parameters,
and the test set to report final results. Here, we 
tune the model hyper-parameters using  
the performance across both NER categories (``Overall'') on the development set. 

 We employed pre-trained 50-dimensional  word vectors from \newcite{Luo:2018}. These pre-trained vectors were derived by training the Word2Vec skip-gram model \cite{Mikolov:2013} on a large text collection of 2 million MEDLINE abstracts.

\newcite{Nils:2017} showed that the BiLSTM-CRF model achieved best performance with 2 BiLSTM layers. Therefore, in our experiment, we only evaluated models up to 2 stacked BiLSTM layers.
The size of LSTM hidden states in each layer was selected from [100, 150, 200, 250]. We achieved the highest F1 score on the development set when using 250-dimensional LSTM hidden states for all models.

By default, each of the additional features (POS, chunking tags, gazetteer match tag) was incorporated into the model via a 10-dimensional embedding. Other hyper parameters were also fixed as in \newcite{Nils:2017} during initialization. 
See tables \ref{config:default} and \ref{config:embedding} for more details. 

In the training process, we used the score on development set to assess model improvement. Early stopping was applied if there was no improvement after 10 epochs. The threshold for a word that was not in the word embedding vocabulary to be added into the embedding was set to 5.  The average training time for each epoch was also recorded. 

\begin{table*}[!t]
\centering
\resizebox{16cm}{!}{
\begin{tabular}{l|lll|lll|lll}
\hline
\multirow{2}{*}{\bf Model}  & \multicolumn{3}{c|}{\bf Chemical  } & \multicolumn{3}{c|}{\bf Disease}  & \multicolumn{3}{c}{\bf Overall } \\
\cline{2-10}
&   P & R & F$_1$ &   P & R & F$_1$&   P & R & F$_1$ \\
\hline
BiLSTM & 87.48 & 91.61 & 89.50 & 78.22 & 83.54 & 80.80 & 83.26 & 87.97 & 85.55 \\
BiLSTM + CNN-char & {90.65} & 90.70 & 90.67 & 79.34 & 82.66 & 80.97 & {85.44} & 87.07 & 86.25 \\
BiLSTM + LSTM-char & 90.47 & {91.64} & \textbf{91.05} & {79.43} & {83.97} & \textbf{81.64} & 85.37 & {88.18} & \textbf{86.76} \\
\hline
BiLSTM-CRF & 90.75  & 90.96 & 90.86 & 80.74 & 83.75 & 82.21 & 86.15 & 87.71 & 86.92 \\
BiLSTM-CRF + CNN-char & 91.64  & {92.24} & \textbf{91.94} & 81.42 & {84.67} & \textbf{83.01} & 86.95 & {88.83} & \textbf{87.88} \\
BiLSTM-CRF + LSTM-char & {92.08} & 91.79 & \textbf{91.94} & {81.48} & 84.22 & 82.83 & {87.20} & 88.38 & 87.79 \\
 \hline
BiLSTM-CRF\textsubscript{+Gazetteer} & 92.26 & 91.01 & 91.63 & 81.87 & 82.19 & 82.03 & 87.53 & 87.03 & 87.28 \\
BiLSTM-CRF\textsubscript{+Gazetteer}+ CNN-char & {92.62} & 92.03 & \textbf{92.32} & 80.72 & {85.28} & \textbf{82.94} & 87.07 & {88.99} & \textbf{88.02} \\
BiLSTM-CRF\textsubscript{+Gazetteer} + LSTM-char & 92.11 & {92.33} & 92.22 & {82.13} & 83.66 & 82.89 & {87.57} & 88.42 & 87.99 \\
\hline
\hline
Att-BiLSTM-CRF (LSTM-char) \cite{Luo:2018} & 92.88 & 91.07 & 91.96& - & - & - & - & - & -  \\
Att-BiLSTM-CRF\textsubscript{POS+Chunking+Gazetteer} (LSTM-char) & {93.49} & {91.68} & \textbf{92.57} & - & - & - & - & - & -\\
\hline 
\hline
TaggerOne \cite{Leaman:2016}  [$\spadesuit$] & {94.2} &  {88.8} & \textbf{91.4} & {85.2} & {80.2} & \textbf{82.6} & - & - & -\\
tmChem \cite{Leaman:2015} [$\spadesuit$] & 93.2 & 84.0 & 88.4 & - & - & - & - & - & - \\
Dnorm \cite{Leaman:2013}  [$\spadesuit$] & - & - & - & 82.0 & 79.5 & 80.7 & - &- & -  \\ 
\hline
\end{tabular}
}
\caption{Results (in \%) on the test set. [$\spadesuit$] denotes results reported on a 950/50 training/development split  rather than our 900/100 split.  As indicated, Att-BiLSTM-CRF 
 used  LSTM-char word embeddings.}% \newcite{Luo:2018} also used the  900/100 split. }
\label{tab:result}
\end{table*}

\section{Main results}

\subsection{Baseline results}

Table~\ref{tab:result} presents our empirical results. The first three rows show the performance of baseline models without the CRF layer, the next three rows show the performance of BiLSTM-CRF models without additional features, and then the next three rows show the results for BiLSTM-CRF models with additional gazetteer features.

As the empirical results in Table~\ref{tab:result} show, the model with CNN character-level embeddings (CNN-char) and the model with LSTM character-level embeddings (LSTM-char) achieved similar overall F1 scores (87.88\% and 87.79\%, respectively),  outperforming BiLSTM-CRF by approximately 1\% in absolute terms. In particular, on chemical NER, both BiLSTM-CRF-based models with character-level word embeddings obtained the same F1 score (91.94\%), while on disease NER the model with CNN-char obtained  slightly higher performance (83.01\%) than the model with LSTM-char (82.83\%). All models with the CRF layer outperformed their respective baseline BiLSTM models in F1 scores for all entity categories.
%The results show that models with CNN-based character-level  embeddings can achieve comparable result to models using LSTM-based character-level  embeddings.

\subsection{Effect of additional features}
When incorporating  additional POS and chunking features into three baseline BiLSTM-CRF-based models, we found that no performance improvement based on the baseline models was observed.

On chemical NER, the additional gazetteer feature improved the baseline BiLSTM-CRF by about 0.8\% while it only improved the baselines BiLSTM-CRF + CNN-char and BiLSTM-CRF + LSTM-char by about 0.3\%, thus clearly indicating that character-level word embeddings can capture unseen word information.  
Considering both NER categories together (``Overall''), the best  performance was also obtained when the gazetteer feature was added, reaching overall F1 scores of 88.02\% and 87.99\%, respectively, for the two CNN-based and LSTM-based character-level embedding models.

\subsection{Comparison with prior work}
The performance comparison between our BiLSTM-CRF-based models and other machine learning approaches to the two studied NER tasks is also shown in Table~\ref{tab:result}. The pattern of chemical NER outperforming disease NER is consistent across all tools.

The Att-BiLSTM-CRF model \cite{Luo:2018} used a BiLSTM-CRF model with LSTM character-level word embedding and an additional attention layer. 
It achieved an F1 score of 91.96\% on chemical NER without additional features. The positive effect of a gazetteer feature was also observed in their results; the model with syntactic and gazetteer features reached an F1 score of 92.57\%. Note that the datasets used in this paper might not be exactly the same as ours due to random sampling.

The last three rows of Table~\ref{tab:result} show the results presented in \newcite{Leaman:2016}, where 950 of the abstracts were used for training and 50 for development (cf.\ our 900/100 split). Dnorm \cite{Leaman:2013} is a model based on pairwise learning to rank on disease name normalization, which achieved F1 score of 80.7\% on disease NER. The tmChem \cite{Leaman:2015} is based on CRF; using numerous hand-crafted features it reached an F1 score of 88.4\% on chemical entities. As a semi-Markov model with a richer set of features for NER tasks, TaggerOne \cite{Leaman:2016} achieved F1 score of 91.4\% and 82.6\% on chemical and disease entities, respectively.

Compared to previous non-deep-learning methods using CRFs, the BiLSTM-CRF models have significant advantage on F1 score of both chemical and disease entities, primarily due to improvement on recall.

\subsection{Discussion}

In our experiment on the effect of additional features, we found that syntactic features such as POS and chunking information did not have clear positive effect on the performance. In contrast, the match/partial match between words and entries in the chemical gazetteer is a good indicator for the presence of chemical entities. Since the Jochem dictionary contains only chemical entities, it is not surprising that the performance on diseases was not substantially impacted by adding the gazetteer feature, although some small variations in performance can be observed, likely due to changed influences from neighboring terms.

The empirical results shown that models  using either CNN-char or LSTM-char  achieve a similar overall F1 score on chemical and disease NER. The results are further comparable with other state-of-the-art models. This indicates that these character-level models have sufficient complexity to learn the generalizable morphological and lexical patterns in biomedical named entity terms. 

On the other hand, as shown by the substantial differences in the number of parameters in Table~\ref{config:embedding},  CNN \cite{LeCun:1989} has the advantage of reduced training complexity as compared to the LSTM models \cite{HochreiterSchmidhuber1997b} under similar experimental settings. In our experimental environment, the execution time of the model with LSTM-char increased 115\% relative to the baseline BiLSTM-CRF model, while it only increased by 25\% for with CNN-char, as detailed in Table \ref{result:time}. Therefore, consistent with prior results on general NER, we conclude that CNN-based embeddings are  preferable to LSTM-based embeddings for BNER.

\begin{table}[t]
\centering
\resizebox{7.5cm}{!}{
\begin{tabular}{|l|p{2.8cm}|l|}
\hline
Model & Avg.\ Runtime per Epoch (seconds) & $\Delta$  \\\hline
BiLSTM-CRF & 106 & 0 \\
\ \ \ \ \  + CNN-char &  \textbf{134} & \textbf{+26\%} \\
\ \ \ \ \ + LSTM-char &  229 & +115\% \\
\hline
 \end{tabular}
}
\caption{Training time of best performing models  (2 BiLSTM layers and 250 LSTM units), computed on a Intel Core i5 2.9 GHz PC.}\label{result:time}
 \end{table}

%In our error analysis, the statistical information for 
We analyzed the error cases of the  CNN-char and LSTM-char models  without additional features: 3326 and 3271 words were incorrectly predicted using CNN-char and LSTM-char, respectively, with  2138 mistakes in common. In errors which only was made by one of the two models, we found that CNN-char made more false positive predictions and fewer false negative predictions, while LSTM-char made approximately an even number of the two kinds of false predictions.

The relationship between the length of words and these errors was also explored. For words less than 20 characters in length, the distribution of errors is almost identical for the two models. However, for longer words, the model with  LSTM-char tends to make more mistakes. This supports prior observations that LSTM can be difficult to apply to long sequences of input \cite{Bradbury:2016}. In approximately 50\% of error cases,  
the word length is short, less than 5 characters. Short biomedical named entities are usually abbreviations and tend to be out-of-vocabulary terms, and are therefore particularly difficult for the character-level word embedding models to capture \cite{Maryam:2017}.

\section{Conclusion}
We compared the performance of BiLSTM-CRF models with CNN-based and LSTM-based character-level word embeddings for biomedical named entity recognition. We confirmed previously published results on chemical and disease NER that demonstrate that  character-level embeddings are helpful.
We further show empirically, generalizing prior results for general NER to the  biomedical context, that there is  little difference between the two approaches:
 both types of  character-level word embeddings achieved identical F1 score on the chemical NER task, and similar performance on disease NER (with CNN-char showing a slight performance advantage). However, the CNN embeddings show a substantial advantage in reduced training complexity.

\section*{Acknowledgments}   
This work was supported by the ARC Discovery Project DP150101550 and ARC Linkage Project LP160101469. 

%\newpage
\bibliography{emnlp2018}
\bibliographystyle{acl_natbib_nourl}

\end{document}